\documentclass[letterpaper]{article}
\usepackage[preprint]{aaai2027}
\usepackage[hyphens]{url}
\usepackage{graphicx}
\usepackage{natbib}
\usepackage{caption}
\usepackage{amsmath,amssymb,amsthm}
\usepackage{booktabs}
\usepackage{algorithm}
\usepackage{algpseudocode}
\usepackage{multirow}

\theoremstyle{plain}

\newtheorem{theorem}{Theorem}
\newtheorem{corollary}{Corollary}

\newcommand{\E}{\operatorname{E}}
\newcommand{\wh}{\widehat}
\newcommand{\TV}{\operatorname{TV}}
\DeclareMathOperator*{\argmin}{arg\,min}

\title{GEAR: Generative Expansion and Real Anchoring for Two-Stage Distillation of Tabular Foundation Models}

\author{
    Qi Qin\textsuperscript{\rm 1,\rm 2}\equalcontrib\thanks{All work performed during internship at Ant Digital Technologies, Ant Group.},
    Jiajie Zhu\textsuperscript{\rm 2}\equalcontrib,
    Dali Chen\textsuperscript{\rm 3},
    Yuzhao Zhang\textsuperscript{\rm 2},
    Jia-Xing Han\textsuperscript{\rm 2},
    Peng Zhang\textsuperscript{\rm 2},\\
    Ying Yan\textsuperscript{\rm 2},
    Yifan Sun\textsuperscript{\rm 1}\corresponding,
    Yu Su\textsuperscript{\rm 2}\corresponding
}
\affiliations{
    \textsuperscript{\rm 1}Renmin University of China, Center for Applied Statistics and School of Statistics, Beijing, China\\
    \textsuperscript{\rm 2}Ant Digital Technologies, Ant Group, Hangzhou, China\\
    \textsuperscript{\rm 3}Nanjing University, Nanjing, China
}

\begin{document}
\maketitle

\begin{abstract}
Tabular foundation models (TFMs) achieve strong performance through in-context learning, but context-dependent inference imposes substantial latency and memory costs, hindering large-scale deployment. We propose GEAR (\emph{Generative Expansion and Real Anchoring}), a modular two-stage framework that distills TFMs into lightweight MLP or tree-based predictors that can be deployed on commodity CPUs. Stage~1 uses synthetic covariates solely as teacher-query locations and trains the student on soft TFM targets, expanding coverage beyond observed rows. Stage~2 re-anchors the student to the target distribution using real labels and out-of-fold teacher predictions, which avoids self-labeling leakage. We further derive a risk certificate characterizing the trade-off between generated-query volume and generator fidelity. Experiments on TALENT and TabArena demonstrate the broad applicability of GEAR. Two-stage MLPs outperform supervised MLPs by 1.81--2.00 AUC points on binary tasks and 1.19--1.35 points on multiclass tasks, with additional gains over real-data-only distillation of 1.76--2.19 and 2.09--2.40 points, respectively. On binary tasks, the gains also transfer to LightGBM and XGBoost, and all three student families outperform CatBoost, the strongest non-TFM baseline, in mean AUC. Ablations show gains beyond longer training or alternative warm starts, greater stability from staged than mixed optimization, and generator-dependent diminishing returns as query volume increases. Finally, GEAR reduces median inference time by 57--2866$\times$ and peak prediction memory by 1.9--3.3$\times$, while retaining higher AUC than matched supervised baselines.

\end{abstract}

\section{Introduction}
Tabular data are central to machine learning in finance, marketing, healthcare, and other high-stakes domains~\citep{grinsztajn2022tree}. Unlike natural language and images, tables are mixed-type, heterogeneous, and dataset-specific, making reusable representations difficult to learn and deploy~\citep{jiang2026representation}. Practical tabular learning has therefore relied on task-specific preprocessing and model selection; gradient-boosted decision trees such as XGBoost~\citep{chen2016xgboost}, LightGBM~\citep{ke2017lightgbm}, and CatBoost~\citep{prokhorenkova2018catboost} remain highly competitive with more complex neural architectures~\citep{van2024tabular}. Recent tabular foundation models (TFMs), including TabPFN~\citep{hollmann2022tabpfn,hollmann2025accurate}, TabICL~\citep{qu2025tabicl,qu2026tabiclv2}, and TabDPT~\citep{ma2024tabdpt}, challenge this paradigm through transformer-based in-context learning (ICL). Given a labeled table as context, they predict query rows without task-specific gradient updates. By learning across large collections of synthetic tabular tasks, these models achieve strong few-shot performance.

\begin{figure}[t]
\centering
\includegraphics[width=0.72\columnwidth]{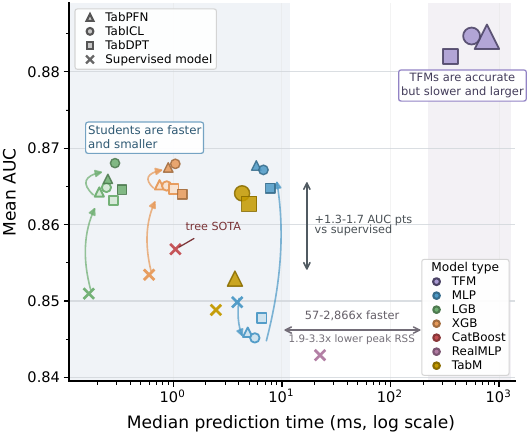}
\caption{Accuracy--efficiency trade-off on TALENT. Mean AUC is plotted against median prediction time; bubble size denotes peak RSS. Shapes denote TFM families, colors denote model types, crosses mark supervised baselines, light fills indicate real-only distillation, and dark fills indicate GEAR. TabM real-only distillation reaches only $\approx0.76$ AUC and is omitted; GEAR students retain teacher accuracy at much lower cost.}
\label{fig:inference-cost}
\end{figure}

This in-context interface, however, creates a deployment bottleneck~\citep{tanna2026pocket,zabergja2026end}. During inference, each query batch requires executing a large teacher while conditioning on the labeled context table, so latency and memory consumption can grow sharply with the context size and query workload. Large-context inference is therefore expensive, prone to out-of-memory failures, and requires training records to remain available at prediction time, raising both efficiency and privacy concerns.

Recent work mitigates this cost by modifying the context used at inference time, including retrieval or nearest-neighbor context selection~\citep{thomas2024retrieval,liu2025tabpfn}, mixture-of-experts routing~\citep{xu2025mixture}, divide-and-conquer inference~\citep{ye2026closer}, and optimal transport-based distillation~\citep{lin2026context}. Although these approaches improve scalability, the predictor still depends on an inference-time context and a large TFM. Consequently, they still incur substantial memory overhead and do not provide the fixed, millisecond-scale latency of a compact, context-free model.

Knowledge distillation (KD) offers a complementary route~\citep{hinton2015distilling,tanna2026pocket}: the teacher's task-adaptive behavior can be amortized into a lightweight MLP or tree-based student. Once distilled, the student has fixed per-row inference cost, runs on commodity CPUs, and no longer requires the labeled context at test time. The main challenge is \emph{query coverage}: the real table provides only finitely many teacher-query locations, which is especially restrictive in small-sample regimes. Moreover, teacher predictions for rows included in the teacher's own context may suffer from self-labeling leakage, motivating out-of-fold~(OOF) real-data anchoring.

\begin{figure}[t]
\centering
\includegraphics[width=0.90\columnwidth]{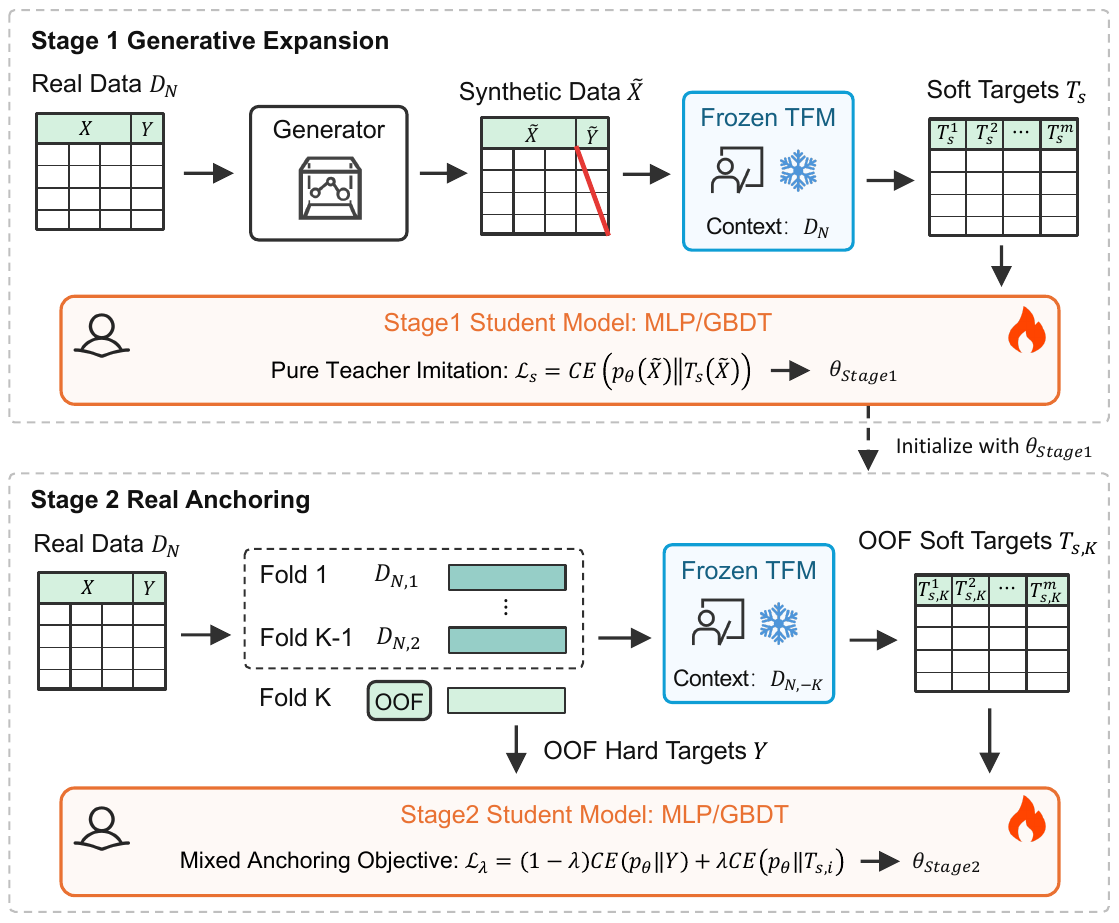}
\caption{Overview of the proposed two-stage generative distillation protocol.}
\label{fig:method-overview}
\end{figure}

To address these challenges, we introduce GEAR, a two-stage protocol that performs \textbf{G}enerative \textbf{E}xp\textbf{A}nsion and \textbf{R}eal Anchoring for in-context TFMs, as summarized in Figure~\ref{fig:method-overview}. Stage~1 queries a frozen teacher on generated covariates and trains the student by teacher-only imitation. Stage~2 initializes from this generated-imitation student and returns to the real table. It uses OOF teacher predictions to avoid self-labeling leakage and optimizes a mixed real-row objective that combines ground-truth labels with teacher soft targets. Thus, synthetic queries expand teacher coverage, whereas real rows provide target-domain anchoring. GEAR is modular and agnostic to generator, teacher, and student architectures, combining diverse tabular generators and TFM teachers with deployable students such as MLPs and tree models.

We evaluate GEAR on TALENT and TabArena with TabICL, TabPFN, and TabDPT teachers, four representative generators, multiple generated-to-real ratios $K$, and MLP or tree-based students. Across teachers, GEAR MLP students outperform supervised MLPs by 1.81--2.00 AUC points on binary tasks and 1.19--1.35 points on multiclass tasks. On binary tasks, Stage~2 improves LightGBM and XGBoost for every teacher, and all three two-stage student families achieve higher mean AUC than CatBoost, the strongest non-TFM baseline. Relative to direct TFM inference, GEAR MLPs retain higher AUC than matched supervised baselines while reducing median prediction time by 57--2866$\times$ and peak prediction memory by 1.9--3.3$\times$ (Figure~\ref{fig:inference-cost}). Volume studies, mechanism ablations, and the risk certificate support the separation of query expansion from real-data anchoring.

Our contributions are summarized as follows:
\begin{enumerate}
\item We introduce GEAR, a modular two-stage framework that combines generative query expansion with real-data anchoring to transfer in-context TFMs into efficient predictors while reducing self-labeling leakage and supporting diverse generator families, teacher models, and student architectures.
\item We conduct extensive experiments showing that GEAR improves over supervised learning and real-data-only distillation, while substantially reducing TFM inference costs and preserving competitive predictive performance.
\item We theoretically link student risk to generated-query volume and fidelity. Our analysis explains the diminishing returns from synthetic expansion, the effects of distribution mismatch, and the role of real-data anchoring in the two-stage design.
\end{enumerate}

\section{Methodology}
\subsection{Setup}
\label{sec:setup}
Let $P$ be the joint distribution of $(X,Y)$ on $\mathcal{X} \times [L]$ with covariate marginal $P_X$, where $X$ is a covariate vector and $Y \in [L]=\{1,\dots,L\}$ is the class label. We observe a training dataset $D_n=\{(X_i,Y_i)\}_{i=1}^n$ drawn independently from $P$. Let $\Delta_{L-1}=\{p\in[0,1]^L:\sum_{a=1}^L p_a=1\}$ denote the probability simplex.

In standard supervised learning, a probabilistic classifier $p_\theta:\mathcal{X}\to\Delta_{L-1}$, parameterized by $\theta\in\Theta$, maps each covariate vector to a predictive distribution. The training data are used only to estimate $\theta$; after fitting, the predictor evaluates new queries independently of the training table. For a supervised loss $\ell:\Delta_{L-1}\times[L]\to\mathbb{R}_+$, such as cross-entropy, the population risk is $R(\theta)=\E_P\{\ell(p_\theta(X),Y)\}$, and empirical risk minimization estimates $\theta$ by minimizing its empirical counterpart.

Tabular foundation models~\citep{hollmann2025accurate} use a different, context-dependent interface. They are pretrained across many synthetic tabular tasks and avoid task-specific parameter updates at deployment. Given a labeled context $C$ and a query $x$, the teacher returns $f_{\mathrm{ICL}}(x\mid C)\in\Delta_{L-1}$. Thus, unlike $p_\theta(x)$, whose predictions are fixed once training is complete, $f_{\mathrm{ICL}}(x\mid C)$ remains coupled to the size, composition, and labels of the context table.

This context dependence creates a practical deployment burden: each prediction invokes the large teacher and requires access to the labeled table. To amortize this cost while preserving the teacher's task-adaptive behavior, we distill the frozen teacher $f_T(x\mid C)=f_{\mathrm{ICL}}(x\mid C)$ into a lightweight student $p_\theta(x)=f_S(x;\theta)$. The student can be instantiated as an MLP, a tree model, or another deployable tabular model.

\subsection{Real-Data Mixed Distillation}
\label{sec:real-kd}
To distill the TFM teacher's behavior into a lightweight parametric student, we use the standard teacher--student formulation with soft teacher targets~\citep{hinton2015distilling}, adapted to tabular TFM distillation~\citep{tanna2026pocket}. For a student distribution $p$ and a teacher distribution $q$ in $\Delta_{L-1}$, the soft-label loss is $\ell_T(p,q)=-\sum_{a=1}^{L}q_a\log p_a$\footnote{This is equivalent to minimizing $\mathrm{KL}(q\|p)$ up to the teacher entropy.}. With mixing weight $\lambda\in[0,1]$, the real-row mixed loss is
\[\ell_\lambda(p, y, q) = (1-\lambda)\ell(p,y) + \lambda\ell_T(p,q).\]

With a single fixed teacher context $C$, the population loss would be $\E_P\ell_\lambda\{p_\theta(X),Y,f_T(X\mid C)\}$. On real training rows, however, asking the teacher to predict a row that is also inside its context can leak that row's label into the distillation target. We therefore use out-of-fold (OOF) teacher predictions, following the real-data TFM distillation baseline of \citet{tanna2026pocket}.

Specifically, we partition $D_n$ into $K_{\rm cf}$ equal folds $I_1,\ldots,I_{K_{\rm cf}}$ of size $m=n/K_{\rm cf}$. For fold $I_k$, let $D_n^{(-k)}=D_n\setminus\{(X_i,Y_i):i\in I_k\}$ and define the fold teacher map $T_k(\cdot)=f_T(\cdot\mid D_n^{(-k)})$. The fold-averaged OOF population loss is
\begin{align}
\label{eq:pop-oof-loss}
   L_{\lambda}^{P}(\theta)
   &=\frac1{K_{\rm cf}}\sum_{k=1}^{K_{\rm cf}}
     \E_{(X,Y)\sim P}
     \ell_\lambda\{p_\theta(X),Y,T_k(X)\}.
\end{align}
The matching empirical OOF objective is
\begin{align}
\label{eq:oof-empirical-loss}
   \widehat L_{\lambda,n}(\theta)
   &=
   \frac1{K_{\rm cf}}\sum_{k=1}^{K_{\rm cf}}
   \frac1m\sum_{i\in I_k}
    \ell_\lambda\{p_\theta(X_i),Y_i,
    T_k(X_i)\}.
\end{align}

The OOF construction prevents self-labeling by ensuring that the teacher target for a real row is produced from a context excluding that row. Unless stated otherwise, all real-data teacher predictions below use this OOF construction. The resulting real-only OOF mixed baseline, which uses hard labels and OOF teacher predictions on real rows only, is
\begin{equation}
\label{eq:output-r}
\wh\theta_R\in\argmin_{\theta\in\Theta}
   \widehat L_{\lambda,n}(\theta).
\end{equation}

\subsection{Synthetic Covariate Augmentation}
\label{sec:generated}
Synthetic data offer a natural means of alleviating data scarcity and can improve learning through pretraining and fine-tuning~\citep{buhler2026causal}. GEAR uses generated rows more conservatively: they provide additional covariate locations for querying the frozen teacher. To avoid bias from generator-provided labels, Stage~1 discards generated labels and uses only generated covariates with teacher soft targets. Concretely, we draw covariates from a tabular generator, such as a GAN or diffusion model~\citep{xu2019modeling,shi2025tabdiff}, and never treat them as labeled target examples.

Let $\widetilde D_M=\{\widetilde X_j\}_{j=1}^M$, where $\widetilde X_j\stackrel{\rm i.i.d.}{\sim}Q_X$ and $Q_X$ denotes the generated covariate law. For generated queries, we write $T_s(x)=f_T(x\mid D_n)$ for the teacher response. The corresponding population and empirical teacher-imitation objectives are
\begin{align}
\label{eq:stage1-pop}
   L_s^{Q_X}(\theta)
   &=\E_{\widetilde X\sim Q_X}
    \ell_T\{p_\theta(\widetilde X),T_s(\widetilde X)\},
    \notag\\
   \wh L_{s,M}^{Q_X}(\theta)
   &=\frac1M\sum_{j=1}^M
   \ell_T\{p_\theta(\widetilde X_j),T_s(\widetilde X_j)\}.
\end{align}
This stage is most useful when generated volume and fidelity are balanced. A larger $M$ provides more teacher-query locations, whereas a low-fidelity $Q_X$ can shift imitation away from the real covariate law $P_X$. Section~\ref{sec:theory} formalizes this trade-off as a finite-query term plus a generator-fidelity floor.

\subsection{Two-Stage Distillation}
\label{sec:twostage}
To exploit broad synthetic coverage while preserving real-data anchoring, GEAR adopts a two-stage protocol. Stage~1 pretrains the student using generated covariates and teacher soft targets. Stage~2 then returns to the real table and optimizes an OOF mixed objective that combines hard labels with OOF teacher predictions, correcting the Stage~1 student toward the target joint distribution $P$. From a transfer-learning perspective, Stage~1 acquires knowledge from auxiliary generated data, whereas Stage~2 performs real-data anchoring in the target domain. Abundant synthetic queries reduce the estimation burden, whereas real samples mitigate distribution shift and align the student with the target distribution~\citep{zhuang2020comprehensive,li2022transfer}.

Given a center $\theta_0$, let $\Theta_2(\theta_0)\subseteq\Theta$ denote the implementation's Stage~2 search region around $\theta_0$. This region may be induced by initialization, early stopping, or regularization. Algorithm~\ref{alg:protocol} summarizes the complete pipeline from generated teacher queries to OOF real-data anchoring.

\begin{algorithm}[t]
\caption{\textbf{Two-stage generative distillation with OOF real-data anchoring.}}
\label{alg:protocol}
\begin{algorithmic}[1]
\Statex \textbf{Input:} real table $D_n=\{(X_i,Y_i)\}_{i=1}^n$, frozen teacher $f_T$, student $f_S(\cdot;\theta)$.
\Statex \textbf{Input:} generated pool size $M$, fold count $K_{\rm cf}$, losses $\ell,\ell_T$, weight $\lambda$, Stage~2 search rule $\Theta_2$.
\Statex \textbf{Output:} two-stage student $\wh\theta_2$.
\Statex \textbf{Stage~1: generated teacher imitation.}
\State Draw generated covariates $\widetilde X_1,\ldots,\widetilde X_M\sim Q_X$.
\For{$j=1,\ldots,M$}
  \State Query $f_T(\widetilde X_j\mid D_n)$.
\EndFor
\State $\wh\theta_1\leftarrow\argmin_{\theta\in\Theta}\wh L_{s,M}^{Q_X}(\theta)$.
\Statex \textbf{Stage~2: OOF real-data anchoring.}
\State Partition $D_n$ into folds $I_1,\ldots,I_{K_{\rm cf}}$.
\For{$k=1,\ldots,K_{\rm cf}$}
  \State Let $D_n^{(-k)}=D_n\setminus\{(X_i,Y_i):i\in I_k\}$.
  \For{each $i\in I_k$}
    \State Query $f_T(X_i\mid D_n^{(-k)})$.
  \EndFor
\EndFor
\State Define $\wh L_{\lambda,n}$ by mixing real-label and OOF teacher terms as in \eqref{eq:oof-empirical-loss}.
\State Initialize at $\wh\theta_1$ and optimize the OOF mixed objective within $\Theta_2(\wh\theta_1)$:
\State \hspace{\algorithmicindent}$\wh\theta_2\leftarrow\argmin_{\theta\in\Theta_2(\wh\theta_1)}\wh L_{\lambda,n}(\theta)$.
\State \textbf{return} $\wh\theta_2$.
\end{algorithmic}
\end{algorithm}

Stage~1 performs pure teacher imitation on generated covariates:
\begin{equation}
\label{eq:stage1-step}
   \wh\theta_1\in\argmin_{\theta\in\Theta}
   \wh L_{s,M}^{Q_X}(\theta).
\end{equation}

Stage~2 starts from $\wh\theta_1$ and returns to the real labeled sample. It optimizes the OOF mixed objective in \eqref{eq:oof-empirical-loss}, anchoring the Stage~1 student to real labels and OOF teacher predictions within the implementation's Stage~2 search region:
\begin{equation}
\label{eq:output-s2}
   \wh\theta_2\in\argmin_{\theta\in\Theta_2(\wh\theta_1)}
    \wh L_{\lambda,n}(\theta).
\end{equation}

The restricted Stage~2 search region does not guarantee a lower empirical OOF loss than an unrestricted real-only fit. Instead, initializing at $\wh\theta_1$ provides a teacher-informed starting point that can facilitate optimization toward a favorable target-domain solution relative to random initialization.

\section{Theoretical Guarantee}
\label{sec:theory}
This section presents a compact risk certificate for the generated-imitation part of GEAR. The baseline is the real-only OOF mixed estimator $\wh\theta_R$ in \eqref{eq:output-r}, and the target criterion is the population risk $R(\theta)=\E_P\{\ell(p_\theta(X),Y)\}$. The certificate characterizes when generated-only teacher imitation can outperform the real-only baseline.

We use the empirical objectives $\wh L_{s,M}^{Q_X}$ and $\wh L_{\lambda,n}$ for synthetic teacher imitation and real-data OOF distillation, respectively. The conversion gaps quantify surrogate-to-target bias. The terms $\epsilon_s$ and $\epsilon_\lambda^{\rm oof}$ are finite-query and cross-fitting radii. Here, $\TV$ denotes total variation, $B\TV(P_X,Q_X)$ is the generator-fidelity floor, and $\underline R_R$ is a conservative lower certificate for the real-only baseline. Formal definitions appear in the supplementary material.

\begin{theorem}[Generated Stage~1 certificate]
\label{thm:generated-beats-real-only}
Under the regularity conditions stated in the supplementary material, with probability at least $1-\delta$,
\begin{align}
\label{eq:stage1-empirical-gap}
    R(\wh\theta_1)-R(\wh\theta_R)
    &\le
    \wh L_{s,M}^{Q_X}(\wh\theta_1)-\wh L_{\lambda,n}(\wh\theta_R)
    \notag\\
    &\quad
    +\epsilon_s(\Theta,M,\delta/2)
    +\epsilon_\lambda^{\rm oof}(\Theta,n,\delta/2)
    \notag\\
    &\quad
    +\mathfrak b_s(\Theta)+\mathfrak b_\lambda(\Theta).
\end{align}
\end{theorem}

When the right-hand side of \eqref{eq:stage1-empirical-gap} is negative, the certificate guarantees that generated-only Stage~1 outperforms the real-only baseline in target risk. The theorem separates two factors that determine whether additional generated rows help. More queries reduce the finite-query price, but they do not remove a persistent mismatch between generated and real covariates. The following specialization makes this volume--fidelity trade-off explicit.

\begin{corollary}[Fidelity--volume specialization]
\label{cor:reflection-stage1}
Suppose the generated-query radius satisfies
\begin{equation}
\label{eq:generated-complexity}
    \epsilon_s(\Theta,M,\delta)
    \le \mathfrak C_s(\Theta,\delta)M^{-1/2}
\end{equation}
for a class-dependent constant $\mathfrak C_s(\Theta,\delta)$.  Using the TV control in the supplementary material, a sufficient condition for $R(\wh\theta_1)<R(\wh\theta_R)$ is
\begin{align}
\label{eq:reflection-stage1-exact}
    &B\TV(P_X,Q_X)
    +2\mathfrak C_s(\Theta,\delta/2)M^{-1/2}
    \notag\\
    &\quad<
    \underline R_R(\delta/2)
    -\inf_{\theta\in\Theta}L_s^{Q_X}(\theta)
    -\beta_s(\Theta).
\end{align}
\end{corollary}

Corollary~\ref{cor:reflection-stage1} isolates the fidelity--volume price
\begin{equation}
\label{eq:fidelity-volume-price}
    H_1(M)=B\TV(P_X,Q_X)
    +2\mathfrak C_s(\Theta,\delta/2)M^{-1/2}.
\end{equation}
The $M^{-1/2}$ term decays with generated volume, whereas $B\TV(P_X,Q_X)$ is an irreducible fidelity floor intrinsic to the generator. When $Q_X$ closely matches $P_X$ so that $\TV(P_X,Q_X)\approx 0$, increasing $M$ progressively tightens the certificate until it reaches this fidelity floor. This yields a scaling law for synthetic expansion: the certifiable finite-query error decays as $M^{-1/2}$ only while sampling error is the primary bottleneck. When $Q_X$ is poorly matched to $P_X$, additional queries cannot mitigate the resulting distribution shift.

Stage~2 addresses the source-induced bias inherited from generated-query imitation. In transfer-learning terms, Stage~1 uses source-like synthetic coverage to reduce the estimation burden and provide an informed initialization, whereas Stage~2 uses real labels and OOF teacher predictions to debias the student under the target distribution. This correction is essential for controlling target risk when $Q_X\neq P_X$~\citep{cortes2014domain,li2022transfer}.

\section{Experiments}
\label{sec:experiments}
\subsection{Experimental Setup}
\label{sec:exp-setup}
We evaluate on the TALENT~\citep{liu2025talent} and TabArena~\citep{erickson2026tabarena} benchmarks. For all settings, we use ROC-AUC for binary classification and macro one-vs-rest AUC for multiclass classification; reported values are averaged over five random runs. The generated-to-real sample ratio for Stage~1 is swept over $K=M/n\in\{1,5,10,20,30,40\}$, where $M$ is the number of generated covariates and $n$ is the real training size.

We use TabICL v2~\citep{qu2026tabiclv2}, TabPFN v3~\citep{grinsztajn2026tabpfn}\footnote{TabPFN v3 is used only for scientific research.}, and TabDPT v1.2~\citep{ma2024tabdpt} as teacher models. The default student is an MLP, and the student-family ablation evaluates LightGBM and XGBoost instead. We additionally benchmark against three strong supervised tabular baselines: CatBoost, RealMLP~\citep{holzmuller2024better}, and TabM~\citep{gorishniy2025tabm}. The primary Stage~1 generator is TabPFGen~\citep{ma2024tabpfgen}, an energy-based in-context tabular generator that conditions on the observed table and produces synthetic covariates without fitting a separate task-specific generator. We also evaluate Copula, CTGAN~\citep{xu2019modeling}, and TabDiff~\citep{shi2025tabdiff}, covering a dependence-preserving statistical generator, a GAN-based generator, and a diffusion-based generator.

We compare the following procedures. The \emph{teacher model} is the frozen TFM evaluated directly with the real training table as context. The \emph{supervised student} is trained only on real data with hard labels. The \emph{synthetic-pretraining student} is trained on generated covariates with soft teacher targets in Stage 1. The \emph{real-data distillation baseline} uses only real rows with out-of-fold teacher predictions and hard labels~\citep{tanna2026pocket}. The \emph{two-stage student (GEAR)} in Section~\ref{sec:twostage} starts from synthetic pretraining and then applies the real-data anchoring step.

\paragraph{Implementation and efficiency.}
The default student is a residual MLP with LayerNorm and SiLU activations, hidden width 512, four blocks, and dropout 0.2; Stage~2 uses a soft-label weight of $\lambda=0.7$. Timing experiments use a server with a single NVIDIA A800-SXM GPU and two Intel Xeon Platinum 8369B CPUs. We time direct TFM inference on the GPU and student inference on the CPU and, where supported, on the same GPU. Detailed optimization settings and implementation are provided in the supplementary material.

\subsection{Main Synthetic-Volume Study}
\label{sec:exp-main}

\paragraph{Q1: Does distillation improve a deployable MLP?}
\begin{figure*}[t]
\centering
\includegraphics[width=0.86\textwidth]{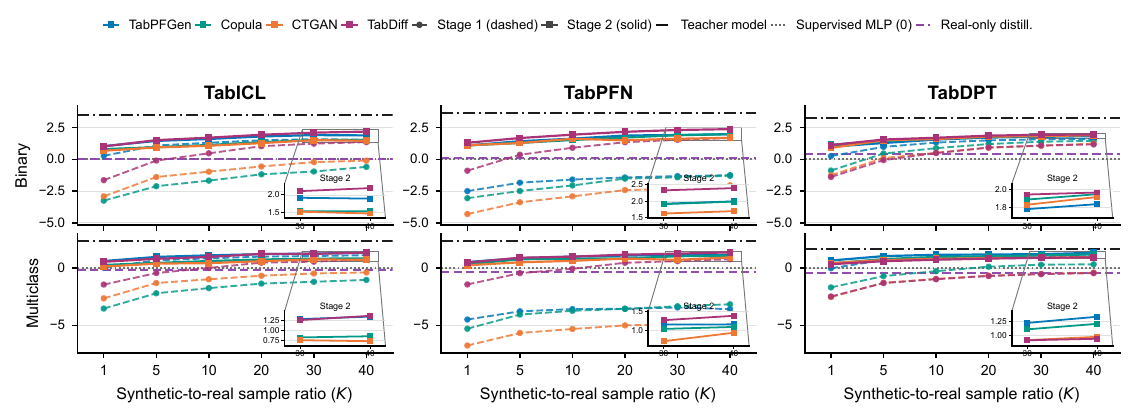}
\caption{Generator trajectories on TALENT. Columns correspond to TabICL, TabPFN, and TabDPT, and rows to binary and multiclass tasks. Curves show mean gains in AUC points over the supervised MLP across generated-to-real ratios $K$. Colors denote generators; dashed circles and solid squares indicate Stage~1 and Stage~2, respectively. Horizontal lines mark the teacher, supervised-MLP, and real-only-distillation references.}
\label{fig:m-sweep}
\end{figure*}

Figure~\ref{fig:m-sweep} compares four generators on matched datasets and reports AUC-point gains relative to the supervised MLP; thus, zero denotes the supervised baseline, whereas the real-data distillation line represents the performance attainable without generated covariates. Across teacher--task panels, the trajectories exhibit a consistent pattern: increasing $K$ improves Stage~1 until the gains plateau at a generator-dependent level, while Stage~2 generally improves the corresponding Stage~1 solution through real-data anchoring.

For TabPFGen, the two-stage MLP at $K=40$ achieves mean AUC gains of 1.91 points on binary tasks and 1.27 points on multiclass tasks over the supervised MLP. As $K$ increases, the marginal benefit of additional generated rows diminishes. From $K=1$ to $K=10$, mean Stage~1 AUC rises by 0.98 points on binary tasks and 0.86 points on multiclass tasks, whereas the increase from $K=10$ to $K=40$ is only 0.29 and 0.17 points, respectively. This saturation is consistent with the volume--fidelity trade-off: additional teacher queries reduce finite-sample error, whereas real-data anchoring is required to mitigate the generated--real distribution mismatch.

\paragraph{Q2~(Stage~1 Analysis): How much does the generator matter?}
Generator choice materially affects both the Stage~1 plateau and the benefit available from Stage~2.

TabPFGen and TabDiff produce richer energy-based and diffusion-based query distributions; their stronger Stage~1 performance is often accompanied by larger two-stage gains. Copula is a useful low-capacity stress test because it preserves a simple dependence structure, but it is not designed to learn complex structural dependencies or causal relationships. CTGAN typically lies between these extremes. As $K$ grows, Stage~1 with higher-fidelity generators can surpass the real-only reference, although weaker generators need more generated rows to reach the same range. By contrast, in the TabICL and TabPFN panels, Copula often remains below real-only distillation even with larger generated pools. This pattern matches the fidelity floor in Corollary~\ref{cor:reflection-stage1}.

Stage~2 usually improves its corresponding Stage~1 curve because it re-optimizes on real rows, but it is not an automatic repair step. If synthetic teacher imitation pushes the student toward a low-performing region, OOF correction on real rows has limited room to recover. Generator quality therefore affects both Stage~1 accuracy and the usefulness of subsequent real-data anchoring. At $K=40$, Table~\ref{tab:catboost-mlp-gain} reports mean paired AUC-point differences and two-sided Wilcoxon $p$-values after averaging available teacher results per task. The fractions of positive paired differences are 57.7--69.8\% (binary) and 60.9--64.4\% (multiclass); TabDiff shows the broadest positive shift, whereas the Copula and CTGAN multiclass means are negative despite positive medians.
\begin{table}[t]
\centering
\small
\renewcommand{\arraystretch}{0.88}
\setlength{\tabcolsep}{0.4mm}
\begin{tabular}{lcccc}
\toprule
Task & TabPFGen & Copula & CTGAN & TabDiff\\
\midrule
Bin.   & $+0.87\,/\,0.04$ & $+0.81\,/\,0.19$ & $+0.71\,/\,0.10$ & $+1.18\,/\,<0.01$\\
Multi. & $+0.26\,/\,0.05$ & $-0.05\,/\,0.08$ & $-0.17\,/\,0.28$ & $+0.18\,/\,0.02$\\
\bottomrule
\end{tabular}
\caption{GEAR MLP versus CatBoost at $K=40$. Cells report mean $\Delta_{\mathrm{GEAR-CB}}$ / two-sided Wilcoxon $p$-value.}
\label{tab:catboost-mlp-gain}
\end{table}

\begin{figure}[t]
\centering
\includegraphics[width=0.9\columnwidth]{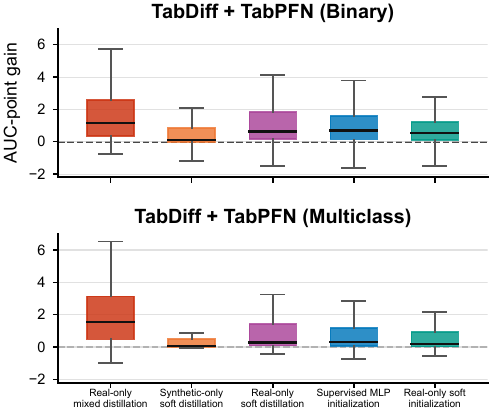}
\caption{Stage~1 replacement ablations using TabPFN as the teacher. Boxplots report paired gains in AUC points of GEAR over five alternatives, ordered from real-only mixed distillation to real-only soft initialization.}
\label{fig:stage1-replacement-representative}
\end{figure}

\paragraph{Q3~(Stage~1 Analysis): Are generated teacher queries really helpful?}
One possibility is that Stage~2 benefits only from the initialization supplied by pretraining or from a longer training schedule. We therefore retain the same two-stage protocol and compare GEAR with five alternatives: \emph{real-only mixed distillation}, \emph{synthetic-only soft distillation}, \emph{real-only soft distillation}, \emph{supervised MLP initialization}, and \emph{real-only soft initialization}. The first three use, respectively, real rows with mixed soft and hard targets, generated rows with soft teacher targets only, and real rows with soft teacher targets only. The final two replace synthetic pretraining with a supervised MLP warm start or a real-only soft-distillation warm start before the same real-data second stage. If synthetic pretraining merely provided an arbitrary initialization or if the gain came solely from a longer schedule, these alternatives should perform comparably. Figure~\ref{fig:stage1-replacement-representative} shows a representative comparison with TabPFN as the teacher. The full result appears in the supplementary material.

The results answer Q3 affirmatively. For TabDiff with a TabPFN teacher on binary tasks, GEAR improves over real-only mixed distillation by 2.61 AUC points and wins on 92.6\% of the paired datasets. It also exceeds supervised MLP initialization by 1.46 points. On multiclass tasks, GEAR gains 2.38 points over real-only mixed distillation and wins on 94.2\% of the paired datasets. Comparisons with supervised MLP initialization and real-only soft initialization show that neither a generic warm start nor additional training explains the gains. Synthetic pretraining yields stronger Stage~2 performance under comparable budgets. Moreover, real-only soft distillation underperforms synthetic-only soft distillation, indicating that generated covariates expand teacher-query coverage rather than merely repeating distillation on observed rows. Thus, the generated stage supplies informative query locations, while the real stage restores target-domain alignment.

\paragraph{Q4: Does GEAR generalize across student families?}
This study replaces the MLP student with LightGBM and XGBoost while retaining the same two-stage logic. The tree-based student study uses TabPFGen as its generator: Stage~1 first trains the tree student on generated teacher targets, and Stage~2 continues training from that Stage~1 student using real-data anchoring. Table~\ref{tab:student-family} separates three paths: Stage~1-only synthetic teacher imitation, real-data distillation from scratch, and the full two-stage route.

\begin{table}[t]
\centering
\small
\setlength{\tabcolsep}{0.3mm}
\begin{tabular}{lllrrr}
\toprule
Teacher & Task & Student & S1 $\Delta$ & Real $\Delta$ & S2 $\Delta$\\
\midrule
\multirow{4}{*}{TabICL} & \multirow{2}{*}{Bin.} & LGBM & $+0.79_{\scriptstyle (0.25)}$ & $+0.92_{\scriptstyle (0.24)}$ & $\mathbf{+1.53}_{\scriptstyle (0.20)}$\\
 & & XGB & $+0.96_{\scriptstyle (0.21)}$ & $+1.11_{\scriptstyle (0.15)}$ & $\mathbf{+1.37}_{\scriptstyle (0.18)}$\\
\cmidrule(lr){2-6}
 & \multirow{2}{*}{Multi.} & LGBM & $-6.34_{\scriptstyle (1.14)}$ & $-8.69_{\scriptstyle (1.04)}$ & $\mathbf{-5.93}_{\scriptstyle (1.14)}$\\
 & & XGB & $+0.22_{\scriptstyle (0.22)}$ & $+0.40_{\scriptstyle (0.13)}$ & $\mathbf{+0.51}_{\scriptstyle (0.12)}$\\
\midrule
\multirow{4}{*}{TabPFN} & \multirow{2}{*}{Bin.} & LGBM & $-2.57_{\scriptstyle (0.76)}$ & $+0.91_{\scriptstyle (0.24)}$ & $\mathbf{+1.32}_{\scriptstyle (0.20)}$\\
 & & XGB & $-2.71_{\scriptstyle (0.90)}$ & $+1.14_{\scriptstyle (0.16)}$ & $\mathbf{+1.38}_{\scriptstyle (0.18)}$\\
\cmidrule(lr){2-6}
 & \multirow{2}{*}{Multi.} & LGBM & $-10.39_{\scriptstyle (1.62)}$ & $-8.96_{\scriptstyle (1.12)}$ & $\mathbf{-6.02}_{\scriptstyle (1.22)}$\\
 & & XGB & $-5.28_{\scriptstyle (1.45)}$ & $+0.15_{\scriptstyle (0.25)}$ & $\mathbf{+0.44}_{\scriptstyle (0.12)}$\\
\midrule
\multirow{4}{*}{TabDPT} & \multirow{2}{*}{Bin.} & LGBM & $+0.90_{\scriptstyle (0.24)}$ & $+0.65_{\scriptstyle (0.27)}$ & $\mathbf{+1.11}_{\scriptstyle (0.23)}$\\
 & & XGB & $+0.69_{\scriptstyle (0.20)}$ & $+0.82_{\scriptstyle (0.20)}$ & $\mathbf{+0.98}_{\scriptstyle (0.18)}$ \\
\cmidrule(lr){2-6}
 & \multirow{2}{*}{Multi.} & LGBM & $-2.50_{\scriptstyle (0.55)}$ & $\mathbf{-1.77}_{\scriptstyle (0.42)}$ & $-2.17_{\scriptstyle (0.48)}$\\
 & & XGB & $-0.36_{\scriptstyle (0.29)}$ & $-0.20_{\scriptstyle (0.26)}$ & $\mathbf{-0.11}_{\scriptstyle (0.22)}$\\
\bottomrule
\end{tabular}
\caption{Tree-student distillation with TabPFGen. Values are paired gains in AUC points over supervised training; subscripts denote standard errors.}
\label{tab:student-family}
\end{table}

Because tree models are non-differentiable and poorly suited to fine-tuning, distillation is challenging; nevertheless, two-stage training improves LightGBM by 1.11--1.53 AUC points and XGBoost by 0.98--1.38 points over supervised training across all three teachers. Aggregated across datasets, these students also surpass CatBoost, the strongest non-TFM baseline, by 0.74 and 0.84 AUC points, respectively, showing that the gains extend beyond matched supervised comparisons. XGBoost is especially stable on multiclass tasks, with positive gains for TabICL and TabPFN and near parity for TabDPT. LightGBM is less stable on multiclass tasks, where all three distillation paths remain below supervised training, consistent with \citet{tanna2026distilling}. Nevertheless, for TabICL and TabPFN, two-stage training is less harmful than real-only distillation. Overall, the benefit is student-dependent. Generated teacher queries can guide a useful learning trajectory, whereas the benefit of real-data anchoring depends strongly on the student family and task type. We also evaluate TabM and MLPs with one to three blocks; shallower MLPs show a larger two-stage advantage. Complete results are provided in the supplementary material.

\subsection{Ablation Studies}
\label{sec:exp-ablation}

\begin{figure}[t]
\centering
\includegraphics[width=0.9\columnwidth]{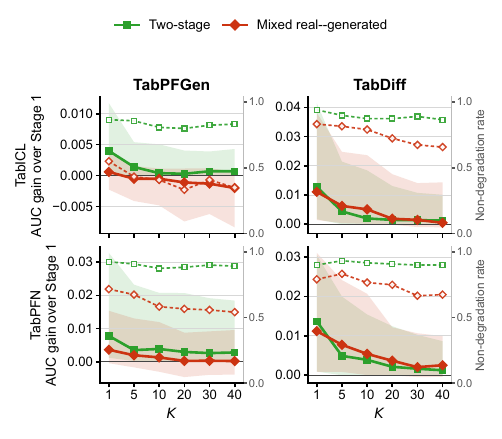}
\caption{Two-stage versus mixed real--generated distillation on TALENT binary tasks. Rows denote TFM teachers, and columns denote TabPFGen and TabDiff. Solid curves show paired median gains in AUC points over Stage~1 with interquartile bands; dashed hollow-marker curves show non-degradation rates relative to Stage~1 (right axis).}
\label{fig:mixed-combined}
\end{figure}

\paragraph{Staged optimization versus mixed training.}
This ablation tests whether the generated and real objectives can be optimized jointly. The mixed real--generated baseline combines generated teacher targets with OOF real-data anchoring but does not separate query expansion from real-data anchoring. Figure~\ref{fig:mixed-combined} reports paired results across $K$ on TALENT binary tasks, including median AUC gains and non-degradation rates relative to Stage~1; each rate is the fraction of paired tasks where the corresponding sequential or mixed method is at least as accurate as Stage~1.

Figure~\ref{fig:mixed-combined} shows representative TabPFGen and TabDiff comparisons; the full sweep appears in the supplementary material. With TabPFGen, two-stage training has higher median gains in AUC points than mixed training for both teachers at every $K$, while the mixed approach approaches or falls below Stage~1 as $K$ grows. Mixed training can be competitive for other generators, but two-stage training achieves non-degeneration rates that are 4.2--48.7 percentage points higher in every comparison. Separating query expansion from real-data anchoring therefore improves stability.

\paragraph{Soft-label mixing.}
We examine GEAR's sensitivity to the weighting parameter $\lambda$, with full results reported in the supplementary material. In Stage~1, pure or near-pure teacher supervision outperforms training with generated hard labels, confirming that generated rows are more useful as teacher-query locations than as labeled examples. In Stage~2, sensitivity to $\lambda$ is weaker but non-negligible: teacher soft targets remain beneficial when balanced with real labels, thereby anchoring the objective to the target distribution.

\section{Related Work}
\label{sec:related}
\paragraph{Tabular foundation models.}
Gradient-boosted trees and tuned neural architectures remain strong tabular baselines~\citep{chen2016xgboost,ke2017lightgbm,gorishniy2021revisiting,grinsztajn2022tree}. TabPFN introduced context-based prediction from a labeled table~\citep{hollmann2022tabpfn}; subsequent TFMs improve pretraining strategies, extend context length, or incorporate more structured-data modeling~\citep{ma2024tabdpt,qu2026tabiclv2,grinsztajn2026tabpfn,wang2026limix}. Scalability methods retrieve, adapt, chunk, or optimize the context supplied to a TFM~\citep{thomas2024retrieval,liu2025tabpfn,sergazinov2025chunked,chen2026vip}.

\paragraph{Distilling tabular foundation models.}
Knowledge distillation compresses powerful predictors into deployable models~\citep{bucilua2006model,hinton2015distilling}. TFM-specific approaches include in-context data distillation, MotherNet, TabDistill, end-to-end compression, real-data-only distillation, and interaction distillation~\citep{ma2024context,muller2025mothernet,dissanayake2025tabdistill,zabergja2026end,tanna2026pocket,jia2026selecting}. GEAR differs by separating generated teacher-query expansion from real-data anchoring, which connects it to sample-bias correction under distribution shift~\citep{cortes2014domain}.

\paragraph{Synthetic tabular data generation.}
Synthetic tabular generators include CTGAN, diffusion models, and TFM-based generators~\citep{xu2019modeling,kotelnikov2023tabddpm,zhang2024mixed,shi2025tabdiff,ma2024tabpfgen,jiang2026tabular}. Their utility depends on generator quality, task type, and evaluation protocol~\citep{shi2025comprehensive,challagundla2025synthetic}; accordingly, GEAR uses synthetic rows as teacher-query locations rather than as labeled target examples.

\section{Conclusion}
\label{sec:conclusion}
We present GEAR, a modular two-stage framework that distills in-context tabular foundation models into efficient, context-free predictors through generative query expansion and real-data anchoring. In the main MLP experiments, GEAR improves over supervised learning and real-data-only distillation while substantially reducing TFM inference costs. Our risk certificate further explains the trade-off between query volume and generator fidelity.

Although this work focuses on classification and single-teacher distillation, GEAR could be extended to regression and multi-teacher distillation. Promising directions include fidelity-aware query generation and combining heterogeneous generators. Another direction is to introduce adversarial distribution alignment during distillation, encouraging generated queries to expand teacher coverage while remaining close to the real-data distribution.

\bibliography{GEAR_AAAI27}

\end{document}